\documentclass[runningheads]{llncs}

\RequirePackage[hyphens]{url} 
\PassOptionsToPackage{hyphens}{url}\usepackage{hyperref} 

\usepackage[T1]{fontenc}
\usepackage{graphicx}
\usepackage{stackengine}
\usepackage{layouts}

\usepackage{array} 
\usepackage{pbox} 
\usepackage{amsmath} 
\usepackage{amssymb}
\usepackage[super]{nth}
\usepackage{marvosym} 
\usepackage{capt-of}
\usepackage{subcaption} 

\usepackage{booktabs}
\usepackage{tabularray}
\usepackage{multirow}
\usepackage{comment}

\usepackage[table]{xcolor}
\usepackage{pgf}
\usepackage{collcell}
\usepackage{layouts}

\newcommand*{\MinNumber}{40}%
\newcommand*{\MaxNumber}{200}%

\newcommand{\ApplyGradient}[1]{%
  \pgfmathsetmacro{\PercentColor}{100*(#1-\MinNumber)/(\MaxNumber-\MinNumber)}%
  \edef\x{\noexpand\cellcolor{black!\PercentColor}}\x\textcolor{black}{#1}%
}
\newcolumntype{R}{>{\collectcell\ApplyGradient}{c}<{\endcollectcell}}

\begin{document}

\title{Subgroup Performance Analysis in Hidden Stratifications}

\author{
    Alceu Bissoto\inst{1,2}\textsuperscript{(\Letter)} \and
    Trung-Dung Hoang\inst{1,2} \and
    Tim Flühmann\inst{1,2} \and
    Susu Sun\inst{3} \and
    Christian F. Baumgartner\inst{3,4} \and
    Lisa M. Koch\inst{1,2}\textsuperscript{(\Letter)}
}
\authorrunning{Bissoto et al.}
\institute{
    Department of Diabetes, Endocrinology, Nutritional Medicine and Metabolism UDEM, Inselspital, Bern University Hospital, University of Bern, Switzerland \and
    Diabetes Center Berne, Switzerland \and
    Cluster of Excellence: Machine Learning - New Perspectives for Science, University of Tübingen, Germany \and
    Faculty of Health Sciences and Medicine, University of Lucerne, Switzerland \\
    \email{\{alceu.bissoto,lisa.koch\}@unibe.ch}
}
    
\maketitle

\setcounter{footnote}{0}
\begin{abstract}
Machine learning (ML) models may suffer from significant performance disparities between patient groups. 
Identifying such disparities by monitoring performance at a granular level is crucial for safely deploying ML to each patient. 
Traditional subgroup analysis based on metadata can expose performance disparities only if the available metadata (e.g., patient sex) sufficiently reflects the main reasons for performance variability, which is not common. 
Subgroup discovery techniques that identify cohesive subgroups based on learned feature representations appear as a potential solution: They could expose hidden stratifications and provide more granular subgroup performance reports. 
However, subgroup discovery is challenging to evaluate even as a standalone task, as ground truth stratification labels do not exist in real data. Subgroup discovery has thus neither been applied nor evaluated for the application of subgroup performance monitoring.
Here, we apply subgroup discovery for performance monitoring in chest x-ray and skin lesion classification.
We propose novel evaluation strategies and show that a simplified subgroup discovery method without access to classification labels or metadata can expose larger performance disparities than traditional metadata-based subgroup analysis.
We provide the first compelling evidence that subgroup discovery can serve as an important tool for comprehensive performance validation and monitoring of trustworthy AI in medicine%
\footnote{Our code is available at \url{https://anonymous.4open.science/r/hidden-subgroup-perf-B560}}.
\keywords{Subgroup discovery \and performance monitoring}
\end{abstract}

\section{Introduction}

Machine learning (ML) models often perform systematically differently across patient subgroups \cite{seyyed2021underdiagnosis,christodoulou2024confidence,koch2024postmarket,oakden2020hiddenStratification}. This has hampered past attempts at safely deploying medical AI in particular in underserved populations~\cite{seyyed2021underdiagnosis,christodoulou2024confidence}.
Model performance can depend on many factors \cite{finlayson2021ClinicianDatasetShift}, including patient attributes (e.g., sex, age, ethnicity) and image characteristics (e.g., image quality, artifacts, device manufacturer).
Subgroup analysis based on such metadata could identify disparate outcomes in patient groups. However, limited metadata typically exists, and available metadata may not adequately capture the data's true variability nor incorporate concepts important to ML models. Hidden stratifications therefore often exist, which can lead to systematic performance disparities that go unnoticed in the evaluation of ML models~\cite{oakden2020hiddenStratification}.

Recently, subgroup discovery methods have emerged for algorithmically identifying systematically different subgroups in computer vision tasks~\cite{d2022spotlight,eyuboglu2022domino,yenamandra2023facts}. These techniques appear as a potential solution for more comprehensive model validation as they could expose hidden stratifications and enable more detailed subgroup performance analyses. 
However, subgroup discovery is challenging to evaluate even as a standalone task, as labels for ground truth stratifications inherently do not exist in real data.
The lack of labels hinders its application to performance monitoring, for which it remains surprisingly underexplored.
As a result, current evaluation approaches are limited to (1) less realistic synthetic datasets, where factors of variations can be fully controlled, or (2) measuring alignment with known characteristics such as patient sex or age, which we realistically cannot expect to characterise the main factors of variation in heterogeneous data distributions.

\begin{figure}[t]
    \centering
    \includegraphics[]{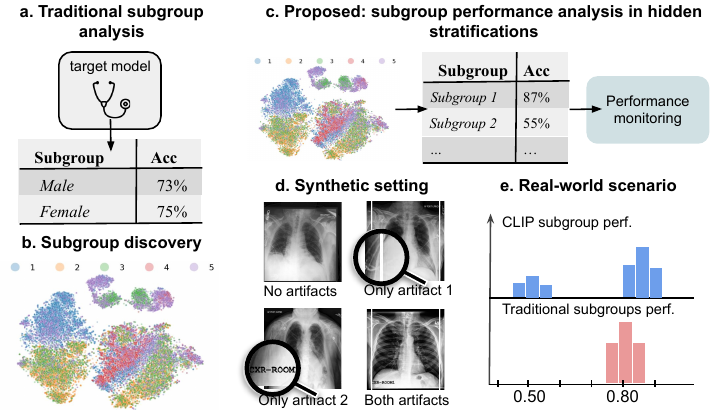}
    \caption{(a) Traditional subgroup analysis detects disparate patient outcomes, but it is limited to annotated metadata. (b) Subgroup discovery reveals hidden stratifications but lacks performance validation. (c) We bridge this gap by applying subgroup discovery for performance analysis in both (d) controlled synthetic settings and (e) real-world scenarios with unknown subgroups.
    }
    \label{fig:graphical-abstract}
\end{figure}

In this paper, we apply and evaluate subgroup discovery in the downstream application of subgroup performance analysis (Fig.~\ref{fig:graphical-abstract}). 
While validation remains challenging, we propose novel evaluation metrics and provide the first compelling evidence on synthetic and real-world medical image classification tasks that subgroup discovery can expose systematic performance gaps. We argue that subgroup discovery can be an effective and easily implemented tool to enhance the performance validation and monitoring of ML systems in medicine.

\section{Methods}
\label{sec:methods}

We seek subgroup divisions that expose large systematic performance gaps of a target classification model while preserving subgroup cohesion, so that the model performance in a subgroup can be attributed to a shared characteristic. Metadata-based subgroups are inherently cohesive since the division is provided by a semantic concept such as patient age or sex. However, we hypothesise that these attributes do not adequately reflect the main factors of variation affecting model performance, which often results in relatively small performance gaps (Fig.\,\ref{fig:graphical-abstract}\,\textbf{a}). Instead, we propose to use subgroup discovery techniques (Fig.\,\ref{fig:graphical-abstract}\,\textbf{b}) for subgroup performance analysis in hidden stratifications (Fig.\,\ref{fig:graphical-abstract}\,\textbf{c}). 

We propose a two-tiered evaluation approach to tackle the difficult challenge of validating hidden subgroups.
First, we inject synthetic artifacts to create clinically-inspired subgroups where ground-truth model performance is available (Fig.\,\ref{fig:graphical-abstract}\,\textbf{d}). 
Finally, we propose a strategy to evaluate subgroup discovery for performance analysis for the first time in a real-world data distribution (Fig.\,\ref{fig:graphical-abstract}\,\textbf{e}). 

\subsection{Preliminaries: Subgroup Discovery Algorithm}

We use DOMINO\,\cite{eyuboglu2022domino}, a simple yet effective approach for subgroup discovery. 
First, a feature representation $z(x)$ is extracted from each image $x$ using an external pretrained model such as CLIP \cite{clip} followed by dimensionality reduction using principal component analysis. In addition, softmax predictions $\hat{y}(x)$ are obtained from the target classification model. 
While the model predictions encapsulate characteristics important for the classification task, the external model helps identify task-agnostic features such as artifacts. Next, the samples are clustered into subgroups $\mathcal{S}$ using a generalised Gaussian Mixture Model (GMM) by minimising the following objective (similar to \cite{eyuboglu2022domino}):

\begin{equation}
    \ell(\phi) \!=\! \!\!\!\!\!\!\!\sum_{i=1}^{n_{samples}}\!\!\!\!\! \log \! \sum_{j=1}^{|\mathcal{S}|} P_{\phi_S}(\mathcal{S}^{(j)}\! =\! 1) P_{\phi_Z}\!(Z\!\! =\!\! z(x_i)\! \mid \!\mathcal{S}^{(j)}\!\! =\!\! 1) P_{\phi_{\hat{Y}}}\!(\hat{Y}\!\! =\!\! \hat{y}(x_i) \!\mid\! \mathcal{S}^{(j)}\! \!=\!\! 1)^\gamma,
    \label{eq:domino}
\end{equation}
\noindent where $\gamma$ balances the influence of predicted labels $\hat{y}(x)$ and embeddings $z(x)$ in the slicing decision. 
In contrast to the original DOMINO \cite{eyuboglu2022domino}, we remove classification labels in the GMM, enabling subgroup discovery in post-deployment scenarios with unlabeled test sets.
We use explicit validation and test sets separations, fitting DOMINO on validation, and inferring subgroups on the test~set. 

\subsection{Synthetic scenario with generated artifacts} \label{sec:synth}

We first evaluate subgroup discovery in a simulated scenario where ground truth subgroups and subgroup performances are known. To simulate performance disparities, we add artifacts spuriously correlated with the positive disease label, similar to standard practice in shortcut learning research \cite{roschewitz2024automatic,bayasi2024biaspruner,bissoto2023even,sun2023right}. 
In particular, we introduce a simulated scenario where we synthetically add two artifacts independently correlated with the label: one is a known attribute for traditional subgroup analysis, but the other is hidden and could potentially be exposed by subgroup discovery. The artifacts are inserted on positive samples with probability, or bias level, $p$, and on negative samples with probability $1-p$, resulting in four ground truth subgroups.
Training and validation sets are generated from this biased version of the data and are used for training the target classification model and selecting its hyperparameters.
The validation set is also used to fit DOMINO.
For testing, we use an unbiased test set ($p=0.5$), facilitating fair comparisons across training bias levels.

\subsection{Real-world data distribution: unknown hidden stratifications}

Next, we assess the ability of subgroup discovery to reveal hidden performance gaps in real-world data where no labels exist for hidden stratifications. 
Following the same procedure as in the synthetic setting, we train the target classification model and DOMINO based on the training and validation set and infer subgroups on the test set.
In the absence of ground truth labels for hidden subgroups, we use measured metadata (e.g. patient age, sex) as a baseline stratification method, which reflects current standard practice for subgroup performance analysis. Each metadata attribute (e.g. patient sex) defines a different subgroup division (male vs. female), assigning each sample its corresponding attribute performance. 
We average the performance values across all its metadata attributes to obtain an overall performance metric for each sample.
For subgroup discovery, we can extend the same idea to marginalize over the stochastic effects caused by the use of different random seeds, providing a more robust estimation of the discovered subgroup performances.

\subsection{Evaluation metrics} 

An ideal stratification leads to subgroups with systematic performance differences. Identifying large performance gaps across cohesive groups may provide actionable insights into the failure modes of the target classification model. 
We propose two new metrics to evaluate the quality of discovered subgroups: performance gap and average purity.  
We measure the \textbf{performance gap} of a subgroup division $S$ as  $\Delta(S) = \max_{s \in S} M(s) - \min_{s \in S} M(s)$, where $M(s)$ is the model performance in subgroup $s$, e.g. accuracy. 
\textbf{Average purity} measures subgroup cohesion by calculating how well subgroups align with known attributes, such as the presence of artifacts or patient characteristics.
For subgroup $s$, let \( n_{s,a} \) be the number of samples with attribute \( a \) and \( n_s \) the total samples. The purity of \( s \) is the fraction of samples in its majority attribute,  corrected by a term $c$ for robustness to small subgroups. Then, the average purity is given by
$AP(S) = \frac{1}{|A|} \sum_{a \in A} \max_{s \in S_a} \left( \frac{n_{s,a}}{n_s + c} \right),$
where $S_a$ is the set of subgroups whose majority attribute is \( a \).

\subsection{Datasets} \label{sec:datasets}
We selected datasets that provide comprehensive coverage of metadata.
CheXpert-Plus~\cite{chambon2024chexpert} is an extension of CheXpert~\cite{irvin2019chexpert} and provides metadata that allows for a challenging comparison to our discovered subgroups. The metadata includes patient demographics (e.g., sex, age), comorbidities (e.g., edema, fracture), 
and artifacts, totalling 20 attributes. Our training, validation, and test set follow an 80/10/10 division, with a total of 178,684 / 22,263 / 22,281 images~respectively.

SLICE-3D~\cite{kurtansky2024slice} is a recent skin lesion classification dataset. Apart from patient details (e.g., sex, age), it includes lesion-specific visual traits, enabling analysis of subgroups aligned with diagnostic-relevant features (e.g., lesion hue and size). Due to the dataset's imbalance, we allocated more samples to the validation and test sets to ensure an adequate number of positive cases. We divided Patient IDs in a 60/20/20 scheme, resulting in 252,047 / 80,516 / 68,496 images.

While we use both datasets for our real-world experiments, we adapt CheXpertPlus with two clinically-inspired artifacts following previous work~\cite{sun2023right} (Fig. \\ \,\ref{fig:graphical-abstract}\,\textbf{b}): a \textit{hospital tag} on the bottom left, and vertical lines of \textit{hyperintense signal}.

\section{Results}

\subsection{Experimental setup}

For all experiments, we trained ResNet-50~\cite{he2016resnet} classification models using SGD and searched over learning rates of $\{10^{-5}, 10^{-4}, 10^{-3}\}$ with the weight decay of $10^{-4}$. Models were selected based on validation balanced accuracy for CheXpertPlus and thresholded AUC for SLICE-3D. For CheXpertPlus, we chose the task of ``cardiomegaly vs. all'', while for SLICE-3D we followed the original problem of ``malignant vs. benign''.
For subgroup discovery, we always used 15 subgroups. The external model is the pretrained CLIP~\cite{clip} for all scenarios, and we included BiomedCLIP~\cite{zhang2024biomedclip} for our real-world CheXpertPlus experiments. 
In the synthetic scenario (Sec.~\ref{sec:synth}), we considered hyperintensities as a known attribute and hospital tags as a hidden stratification and varied the bias level $p$ between {0.6, 0.7, 0.8}.
We use accuracy as our primary metric for measuring performance gaps and subgroup performances.

\subsection{Subgroup discovery uncovers large performance disparities while maintaining cohesive subgroups }
\label{sec:hparam-level2}

\begin{figure}[t]
    \centering
    \includegraphics[width=\linewidth]{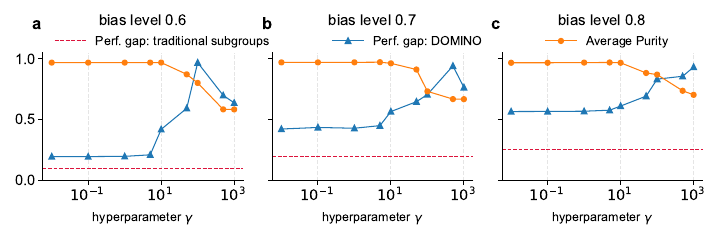}
    \caption{Performance gap and  purity of subgroups across different $\gamma$ and bias levels.  
    }
    \label{fig:l2-ablation}
\end{figure}
Across all experiments in both synthetic (Fig.\,\ref{fig:l2-ablation}) and real-world settings (Fig.\,\ref{fig:l3_cxr}\,\textbf{a, c}), subgroup discovery consistently exposed performance gaps larger than traditional subgroups (red dash line in Figs.\,\ref{fig:l2-ablation}\,\textbf{a-c}) without sacrificing cohesion. While the performance gap and purity competed, performance gaps increased before purity declined when increasing $\gamma$. This allowed substantial performance disparities to be exposed without sacrificing the cohesiveness of the subgroups.
We chose the ``elbow'' point before a sharp purity decrease, resulting in $\gamma=10$ for synthetic and real-world CheXpertPlus, and $\gamma=50$ for the SLICE-3D, as shown in Figs.\,\ref{fig:l2-ablation} and \ref{fig:l3_cxr}\,\textbf{a, c}.

\subsection{Subgroup discovery captures actual subgroup performance} \label{sec:results_known_biases}

\begin{figure}[t]
    \centering
    \includegraphics[]{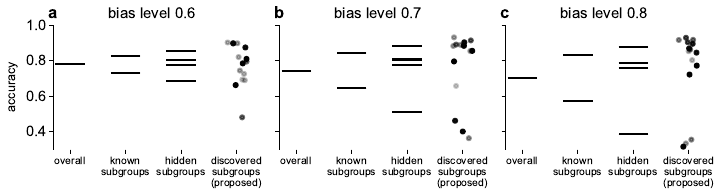}
    \caption{Detailed subgroup accuracies for our synthetic scenario. Purer subgroups performances (darker dots) capture the true performance gap characterized by hidden subgroups, which are overlooked by traditional subgroup analysis with access to a single artifact (known subgroups), and by overall performance.
    }
    \label{fig:l2}
\end{figure}

In our synthetic scenario with one known and one unknown artifact, subgroup analysis based on the known artifact unsurprisingly revealed increasing performance gaps when increasing the bias level from 0.6 to 0.8, but missed the much larger, hidden performance gaps caused by the second artifact (Fig.\,\ref{fig:l2}\,\textbf{a-c}).
Subgroup discovery without access to either artifact annotations successfully found subgroups that captured the hidden subgroup performances (dots in Fig.\,\ref{fig:l2}\,\textbf{a-c}). 

As our simulated artifacts were added to real data where factors unknown to us could additionally affect performance, discovered subgroups exposed additional performance disparities. For example, in Fig.\,\ref{fig:l2}\,\textbf{b}, one discovered subgroup neither aligned with the hidden subgroups in terms of performance, nor in terms of purity (reflected by light grey colour). 

\subsection{Subgroup discovery exposes higher performance gaps than traditional subgroup analysis in real-world scenarios} \label{sec:unknown_biases}

\begin{figure}[t]
    \centering
    \includegraphics[]{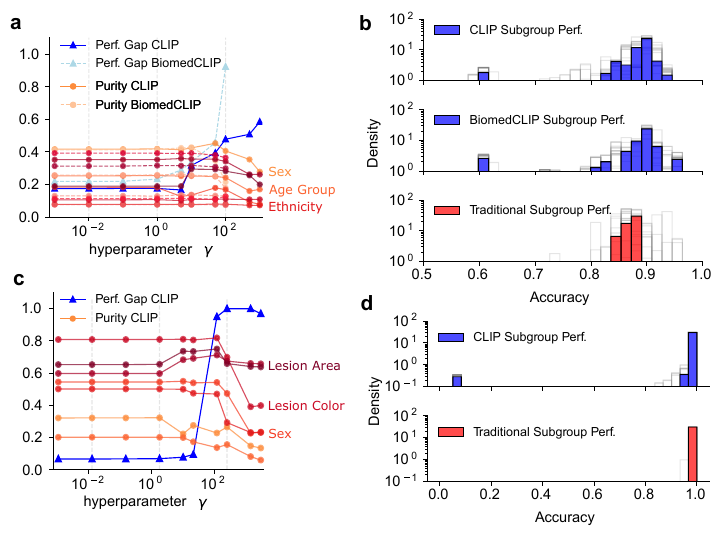}
    \caption{(a,c): Performance gaps and metadata-based purity for different $\gamma$. (b,d): Histograms of subgroup performances for different subgroup divisions: In blue, subgroup discovery with different external CLIP models, averaged over different random seeds (gray transparent bars). In red, our baseline of subgroups defined by different metadata, averaged over their attributes (gray). Top (a,b) and bottom (c,d) rows show CheXpertPlus and SLICE-3D results, respectively.}
    \label{fig:l3_cxr}
\end{figure}

Finally, we applied subgroup discovery for performance analysis in two real-world applications without artificial artifacts in chest x-ray and skin lesion analysis. 
In both cases, hidden biases were likely present but not annotated~\cite{jimenez2023detecting,bissoto2020debiasing}. 
Subgroup discovery identified higher performance gaps than traditional metadata-based analysis (see performance histograms in Fig.\,\ref{fig:l3_cxr}\,\textbf{b, d}).  
On CheXpertPlus (Fig.~\ref{fig:l3_cxr}\,\textbf{b}), subgroup discovery consistently found underperforming subgroups with less than 60\% accuracy, while the majority of subgroups achieved around 90\% accuracy. In contrast, metadata-based analysis did not expose such low-performing subgroups and led to a narrower range of performances overall.
For skin lesion analysis, subgroup discovery found a subgroup with 721 negatives and 17 positives with only 5\% accuracy (Fig.~\ref{fig:l3_cxr}\,\textbf{d}).

\subsection{Discovered subgroups in real-world scenarios do not capture patient demographics, but align well with visual features} \label{sec:res_align}

In the CheXpertPlus dataset, the discovered subgroups did not align well with concepts described by the available metadata, leading to subgroups with low purity concerning attributes such as patient sex, age or ethnicity (Fig.\,\ref{fig:l3_cxr}\,\textbf{a}). This confirms that available metadata often does not reflect the main factors of variability in real-world data distributions.

In contrast, the SLICE-3D skin lesion dataset contained annotations of visual features such as lesion area or colour. The discovered subgroups were well stratified by these visual features. This was reflected by high purity across a wide range of DOMINO configurations~(Fig.\,\ref{fig:l3_cxr}\,\textbf{c}). Demographic attributes such as patient sex remained at a low purity level, similar as in the CheXpert experiments. While some annotated lesion characteristics (e.g. area, colour) are related to lesion malignancy~\cite{abbasi2004early}, subgroup analysis based on these attributes did not expose the performance disparities we observed with discovered subgroups.

\subsection{Feature extractors trained on natural images are sufficient for exposing meaningful performance gaps}

Finally, we used BiomedCLIP~\cite{zhang2024biomedclip} as a feature extractor for subgroup discovery in CheXpert to investigate whether representations learned from biomedical data led to better stratification of disease-related features in medical images. However, BiomedCLIP and original CLIP led to similar subgroup purities (Fig.\,\ref{fig:l3_cxr}\,\textbf{a}) and performance disparities (Fig.\,\ref{fig:l3_cxr}\,\textbf{a, b}). This indicates that even feature extractors trained on natural images can expose meaningful performance gaps in real-world data distributions, where factors of variation may be more visually subtle than the simulated artifacts we introduced in our synthetic experiments.

\section{Discussion}

We demonstrate that hidden stratifications in synthetic and real-world data can lead to performance disparities, which often cannot be detected by traditional metadata-based subgroup analysis. Meanwhile, subgroup discovery exposed substantial and systematic performance disparities between cohesive subgroups.
In the synthetic scenario, discovered subgroups accurately captured artificial ground truth subgroups (Sec.~\ref{sec:results_known_biases}).
In real-world data, where the true factors of variation in data might be more visually subtle, we showed evidence that the factors guiding subgroup discovery are not necessarily low-level perceptual features (Sec.~\ref{sec:res_align}). For skin lesion analysis, the lesion color, which is clinically relevant for the diagnosis of melanoma, indirectly influenced the subgroup discovery, resulting in high average purity.
While no ground truth stratification labels exist for real data, our results were robust and consistent across datasets, hyperparameter configurations and random seeds. 
We conclude that subgroup discovery should be highly relevant as a performance monitoring and reporting tool, and argue that it should accompany traditional subgroup analysis as an additional safeguard during real-world ML validation and deployment.

Future work could further investigate subgroup discovery robustness, facilitating their adoption by ML practitioners. 
Beyond their use in safe deployment, our subgroup performance analysis approach could be useful for developing unbiased ML models. There, discovered subgroups could replace or augment existing subgroup labels, e.g. when reporting worst-group performance.

\bibliographystyle{splncs04_without_urls}
\bibliography{references.bib}

\end{document}